\documentclass{article}
\usepackage{PRIMEarxiv}
\usepackage[utf8]{inputenc}
\usepackage[T1]{fontenc}
\usepackage{hyperref}
\usepackage{url}
\usepackage{booktabs}
\usepackage{amsfonts}
\usepackage{nicefrac}
\usepackage{amsmath}
\usepackage{microtype}
\usepackage{lipsum}
\usepackage{fancyhdr}
\usepackage{graphicx}
\graphicspath{{media/}}
\title{Recent Deep Semi-supervised Learning Approaches and Related Works}

\author{
  Gyeongho Kim \\
  Department of Industrial Engineering \\
  Ulsan National Institute of Science and Technology (UNIST) \\
  \texttt{kkh0608@unist.ac.kr} \\
}

\begin{document}
\maketitle

\begin{abstract}
This work proposes an overview of the recent semi-supervised learning approaches and related works. Despite the remarkable success of neural networks in various applications, there exist a few formidable constraints, including the need for a large amount of labeled data. Therefore, semi-supervised learning, which is a learning scheme in which scarce labels and a larger amount of unlabeled data are utilized to train models (e.g., deep neural networks), is getting more important. Based on the key assumptions of semi-supervised learning, which are the manifold assumption, cluster assumption, and continuity assumption, the work reviews the recent semi-supervised learning approaches. In particular, the methods in regard to using deep neural networks in a semi-supervised learning setting are primarily discussed. In addition, the existing works are first classified based on the underlying idea and explained, then the holistic approaches that unify the aforementioned ideas are detailed.
\end{abstract}

\section{Introduction}

\subsection{Deep Learning}

Since the revival of neural networks with the advent of the AlexNet \cite{krizhevsky2012imagenet}, so-called deep learning approaches employing neural networks having deep architecture (10 or more layers) have been dominating the state-of-the-art methods in various fields \cite{schmidhuber2015deep}. Not only confined to the computer vision field, but they have also shown useful and become a de facto solution in other fields, such as natural language processing, speech recognition, drug discovery, and manufacturing \cite{schmidhuber2015deep,lecun2015deep,goodfellow2016deep,kim2024developing}. Despite the pronounced performance of neural networks, several realistic constraints exist, such as the need for a large amount of labeled data for training to ensure high performances, intensified due to the increasing complexity of neural networks. In fact, most of the recent deep neural network models have been trained and validated on large labeled datasets, such as ILSVRC-2012 (i.e., ImageNet), CIFAR, and MNIST datasets. When there are more unlabeled data than the labeled ones, it is, hence, inevitable to consider using alternative training methods. In this work, the author details a useful deep learning technique, \emph{semi-supervised learning}, which is applicable in scarce-label circumstances.

\subsection{Semi-supervised Learning}

Semi-supervised learning is a setting in which a machine learning model is trained with labeled and unlabeled data simultaneously \cite{grandvalet2005semi}. The existing studies aim to incorporate unlabeled data to enhance the performance of models by fully leveraging the potentially beneficial information given by unlabeled data \cite{chapelle2009semi,oliver2018realistic}. Since human annotations required to generate labels for data are expensive and time-consuming, semi-supervised learning can be a helpful alternative. Prior to broad applications of deep neural networks, conventional machine learning algorithms (e.g., mixture models, support vector machines) have been widely used, coupled with self-training and co-training schemes with confidence-based filtering \cite{zhu2005semi,sajjadi2016regularization,kim2023developing}, but are not discussed in detail in this work.

\section{Deep Semi-supervised Learning}

The following subsections encompass a variety of approaches that have been employed in semi-supervised learning with deep neural networks. Most of them are intertwined and share the same underpinnings, such as the basic assumptions of semi-supervised learning. In addition, the labeled data refer to the input coupled with respective labels $D_{L}=((X_{1},y_{1}),\dots,(X_{l},y_{l}))$ and the unlabeled data that lack labels $D_{U}=(X_{1},\dots,X_{u})$, where normally $u$ >{}> $l$ holds. To reiterate, how to utilize unlabeled data so as to improve supervised neural networks is the main purpose of deep semi-supervised learning.

\subsection{Basic Assumptions}

There exist three basic assumptions in semi-supervised learning, which are detailed as follows. All of the assumptions are basic complementary principles that are used as the underlying bases for semi-supervised learning studies. \\
\textbf{Manifold assumption} First of all, \textit{manifold assumption} states that the data, especially having a high dimensionality lie in a lower-dimensional space called a \textit{manifold}. Based on this assumption, semi-supervised learning is expected to better learn the manifold using both labeled and unlabeled data. In more detail, unlabeled data would help find the data manifold, leading to a better estimate of the decision boundary \cite{oliver2018realistic}. \\
\textbf{Continuity assumption} Secondly, \textit{continuity assumption}, as its name tells, states that when data points are close to each other in a certain space, they are likely to have the same class (e.g., label). This assumption can also be intertwined with the manifold assumption, by applying the same agenda to other dimensional spaces. \\
\textbf{Cluster assumption} Lastly, \textit{cluster assumption} is an extension of the previously stated \textit{continuity assumption}. It states that if data instances belong to the homogeneous cluster, they are likely to have the same class (e.g., label) \cite{verma2019interpolation}. In addition, this assumption implies that the decision boundaries separating different clusters must lie in low-density regions.

\subsection{Entropy Regularization}

One of the most fundamental directions of training neural networks in a semi-supervised setting is to apply an entropy regularization method \cite{grandvalet2005semi}. When a softmax function is used as a final output function for a neural network, the output values are considered as predicted probabilities of each class provided by the model. The entropy regularization leads to a minimization of entropy of the prediction on unlabeled data, thus enforcing the neural networks to output confident predictions for unlabeled data, regardless of the ground truth labels. Minimizing the entropy of the predictions also helps to place decision boundaries in a low-density area, intuitively due to the promotion of confident predictions of models. Based on the principle of entropy minimization, many of the current semi-supervised learning methods utilize a self-training scheme. Using pseudo-labels, which are the virtual labels assigned to the unlabeled data during training, is one of a kind \cite{lee2013pseudo}. Setting the most confidently predicted class as a pseudo-label for unlabeled data and applying the same loss function (e.g., cross-entropy) on a virtual label in a supervised fashion during training lead to better generalization performance as well as a more condensed prediction of data. Despite its simplicity, the pseudo-label is one of the most widely used semi-supervised learning techniques. As opposed to entropy minimization, there also exists a conceptually contrary scheme in which the output distribution of a neural network is penalized \cite{pereyra2017regularizing}. Similar to label smoothing, which has proven effective in generalization \cite{szegedy2016rethinking}, regularizing the output of a model to have higher entropy than a threshold is suggested. However, minimizing the entropy of model prediction is still considered the de facto approach.

\subsection{Self-training}

Self-training in semi-supervised learning literature denotes a learning setting where the supervision on unlabeled data is given by the own prediction of the model trained with the labeled data \cite{zhu2005semi,rosenberg2005semi}; thus, literally, the model is \textit{training itself}. Subsuming the concept of the pseudo-label \cite{lee2013pseudo}, a model trained on the fraction of data is used to generate targets for unlabeled data. Laine and Aila suggest a conceptually similar approach called self-ensembling, including $\Pi$-model and Temporal Ensembling \cite{laine2016temporal}. $\Pi$-model applies stochasticity by evaluating the same network twice with different perturbations (e.g., stochastic augmentation, Gaussian noise, dropout) in a single epoch to force consistent predictions by the stochastic networks. The Temporal Ensembling has the same structure as the $\Pi$-model but uses predictions of previous networks from earlier training stages for ensembling. In both implementations of self-ensembling, ordinary supervised loss for labeled data is combined with a time-dependent weighted unsupervised loss for training. Mean Teacher further improves the approach of Temporal Ensembling, suggesting averaging the model weights instead of predictions, with a concept of Mean Teacher that consists of previous student models from earlier training stages \cite{tarvainen2017mean}. Based on the aforementioned approaches, NoisyStudent further develops self-training for semi-supervised learning by employing an iterative training scheme using noisy student models with stochasticity via stochastic depth, dropout, and RandAugment \cite{xie2020self}. The teacher model generates pseudo-labels on unlabeled data, and a larger noisy student model is trained on both labeled and unlabeled data. The newly trained student model becomes a new teacher model, and the procedure is repeated. With this self-training scheme, NoisyStudent has been shown to provide more stable predictions.

\subsection{Consistency Regularization}

A number of studies exploit consistency regularization based on \textit{cluster assumption}, which states that the data points in the same cluster are likely to have the same class, and those having other classes can be separated by a decision boundary placed on the low-density region. The underpinning is that, since the instances on the same class are close in distance, the model prediction should be invariant to a sensible amount of perturbation on data \cite{athiwaratkun2018there}. Applying consistency regularization in a semi-supervised learning setting is especially focused on regularizing the predictions of neural networks on unlabeled data. Expecting models to make consistent predictions even on unlabeled data would lead it to spontaneously derive useful information of the underlying data distribution from the unlabeled data. An important consideration is the type and the degree of perturbation. In other words, how to effectively noise unlabeled data so as to eventually make the model more robust while pushing the decision boundary towards low-density regions is a matter of issue. Theoretically, any kind of local perturbation, such as Gaussian noise could be helpful in a semi-supervised learning setting \cite{bishop1995training,miyato2018virtual}, if it could function as a regularization for robust training. Consistency regularization can also be imposed by giving stochasticity to models. For instance, using dropout \cite{srivastava2014dropout} or stochastic depth \cite{huang2016deep} to give random noise in model-level also proves effective, such as pseudo-ensemble \cite{bachman2014learning}. The Ladder network \cite{rasmus2015semi} is also aligned with the same idea. A neural network resembling a denoising autoencoder with layers that denoise intermediate representations is trained, combining supervised and unsupervised learning schemes. The stacked layers express rich representations in a hierarchical scheme and prove useful in the semi-supervised learning setting. Using different variations of linear and nonlinear perturbation, such as randomized augmentation, dropout \cite{srivastava2014dropout}, and random pooling has shown effectiveness when the model is expected to output the consistent prediction on perturbed instances \cite{sajjadi2016regularization}. Consistency regularization could be implemented through data-dependent approaches, such as data augmentation and data-agnostic approaches, which are discussed thoroughly in the next subsections.

\subsubsection{Data-dependent Augmentation-based}

Most of the recent works on semi-supervised learning exploit various data augmentation techniques with the intention of regularizing the predictions of the model as a way of consistency regularization. In addition to relatively simple augmentation methods, such as random cropping, flipping, and rotating, more elaborate methods, such as AutoAugment \cite{cubuk2019autoaugment}, RandAugment \cite{cubuk2020randaugment}, and Shake-Shake regularization \cite{gastaldi2017shake} are widely used in semi-supervised learning in order to endow an invariance to the perturbations especially on unlabeled data. In particular, data augmentation can be viewed as a suitable regularization method for neural networks. In fact, rather than using random noise (e.g., Gaussian noise, dropout noise), encouraging a consistency in predictions with several augmentation techniques in an unsupervised manner has shown improvement of performances of neural networks in the vision and natural language processing domain \cite{xie2020self}. A more sophisticated way of determining a direction of data perturbation is the VAT (Virtual Adversarial Training), which stems from the idea of adversarial training \cite{goodfellow2014explaining}. In VAT, the input is perturbed in a way that distracts the model prediction the most, in other words, which makes the prediction most different by applying any kind of perturbation \cite{miyato2018virtual}. Using the current inferred output as a virtual label to compute the adversarial direction and minimize the difference between the predictions with and without perturbation, the suggested method proves effective in practice.

\subsubsection{Data-agnostic Continuity-based}

A large number of studies on consistency regularization make use of various perturbations (e.g., data augmentation) that are data-dependent. In fact, some of the data-dependent augmentation methods are confined to be applied only to image data. However, there also exist several studies that use multiple data instances to enforce consistency, thus considering the vicinity relation between data instances of different categories, called the MixUp \cite{zhang2017mixup}. MixUp generates virtual synthetic data with a convex combination (i.e., linear interpolation) of two data instances also with their labels, assuming that the features, as well as the labels, would have a linear relationship. Enforcing this strict linearity is expected to guide a model to behave linearly in between the data instances and to be robust to adversarial examples \cite{zhang2017mixup,thulasidasan2019mixup}. Verma et al. also extend the idea by encouraging a model to yield consistent predictions on the interpolation of two instances from the unlabeled data \cite{verma2019interpolation}. Imposing the interpolation-based method is a type of consistency regularization imposed on unlabeled data. However, Guo et al. present an intrinsic problem of the MixUp, called the manifold intrusion, which is that some type of interpolated instances collide with the existing data \cite{guo2019mixup}. In order to improve the MixUp, an adaptive version named AdaMixUp is proposed that allows the higher-fold (more than two) MixUp operations \cite{guo2019mixup}. 
Verma et al. also further improve the idea of MixUp by performing the operation on the intermediate representations of data instances at layers in a neural network that are chosen randomly. It is expressed as a manifold MixUp since the MixUp procedure is conducted on any generalized manifold so as to lead to a smoother decision boundary that would be located far away from the data instances \cite{verma2019manifold}.
Compared to data augmentation-based methods, MixUp is easy to implement and data-agnostic, meaning that it can be applied to any type of data.

In a nutshell, the consistency regularization approaches of semi-supervised learning aim to regularize model predictions to be consistent even when the data are perturbed (with augmentation or noise), either in input space or intermediate representation space. As a result, the decision boundaries are set far away from high-density regions. Furthermore, since the regularization is also applied to unlabeled data, the decision boundaries are expected to be closer to the actual low-density regions.

\subsection{Self-supervised}

Self-supervised learning is a setting where a model is trained for various pretext (surrogate) tasks that use targets generated by the data itself without manual annotations. Due to an inherent learning condition, self-supervised learning is used to improve the richness of data representation by neural networks as well as to achieve higher-level visual semantic learning \cite{misra2020self,zhai2019s4l}. In this regard, self-supervised learning is closely connected with semi-supervised learning, since both aim to benefit from unlabeled data in terms of better representation learning and making the most use of unlabeled data \cite{hendrycks2019using}. In fact, self-supervised learning methods have proven effective in a semi-supervised setting where only a small fraction (e.g., 1\% and 10\%) of data are labeled \cite{misra2020self}. Zhai et al. propose adopting self-supervised pretext tasks, such as image rotation prediction and exemplar methods (e.g., various transformations) in the semi-supervised learning scheme, dubbed the $S^{4}L$ (self-supervised semi-supervised learning) \cite{zhai2019s4l}. When self-supervised methods, which encourage neural networks to learn diverse and robust representations, are employed in a semi-supervised learning setting, the models show improved performances. In addition, when multiple semi-supervised learning approaches, such as VAT \cite{miyato2018virtual}, entropy minimization \cite{grandvalet2005semi}, and pseudo-labeling \cite{lee2013pseudo} are used along with self-supervised methods, it even outperformed a fully supervised learning approach, thus suggesting that the semi and self-supervised learning are in fact complementary approaches. In the same sense, Chen et al. propose using deep neural networks in a consecutive training scheme, which consists of self-supervised (unsupervised) task-agnostic pretraining, fine-tuning, and task-specific self-training distillation for semi-supervised learning \cite{chen2020big}. In both the pretraining (i.e., task-agnostic) and distillation (i.e., task-specific) phases, the unlabeled data are utilized. It is worth noting that using unlabeled data largely improved the performance, especially in the self-distillation phase, implying that the unlabeled data can enhance the richness of data representation. Considering that there exist a large number of researches on self-supervised learning, semi-supervised learning could elicit advantages, especially using various pretext tasks that help to learn useful diversified feature representation.

\section{Holistic Approaches}

Since the late 2010s on the frontier of semi-supervised learning, recent works embrace the aforementioned approaches, such as MixMatch, ReMixMatch, and FixMatch \cite{berthelot2019mixmatch,berthelot2019remixmatch,sohn2020fixmatch}. Berthelot et al. suggest MixMatch that employs data augmentation, MixUp \cite{zhang2017mixup}, and soft (i.e., temperature-sharpened) pseudo-labels as guessed labels in a unified manner \cite{berthelot2019mixmatch}. Temperature scaling is also used for sharpening the output of the model in order to reduce the entropy of the distribution. Berthelot et al. suggest a unified approach, dubbed ReMixMatch, which further improves MixMatch, by additionally introducing techniques called distribution alignment and augmentation anchoring \cite{berthelot2019remixmatch}. The distribution alignment makes the distribution of the prediction of a model on unlabeled data match that on the labeled data. The augmentation anchoring suggests generating an anchor target with the prediction on the weakly augmented data and using it as a target for consistency regularization methods with strong augmentations. FixMatch suggests a conceptually similar approach in the sense that it also uses a prediction on weakly augmented unlabeled data as an artificial pseudo-label \cite{sohn2020fixmatch}. In addition, the thresholding on the confidence of the prediction to include as a pseudo-label naturally encourages the minimization of the entropy of the model predictions. Based on the premises of semi-supervised learning, recent approaches focus on how to further improve by unifying the various techniques, especially applicable to deep neural networks.

\section{Conclusion and Discussion}

Semi-supervised learning is an efficient and effective learning scheme in scarce-label circumstances where abundant unlabeled data exist. It is worth reiterating that the primary goal of semi-supervised learning is to well incorporate unlabeled data so as to improve the performance of the model. Recent approaches in semi-supervised learning broadly utilize the aforementioned concepts, such as entropy minimization, consistency regularization, and self-training, through diverse ways of implementation. Most of the methods are not only compelling but also readily applicable in various tasks where deep neural networks are employed. Despite given the appropriate implementation details, such as preprocessing, initialization, augmentation, and regularization, when unlabeled data contain out-of-distribution data, in other words, data having different distribution from that used during training, the performance of semi-supervised learning methods degrades significantly \cite{oliver2018realistic}. Thus, the assumption that unlabeled data and labeled data share the same underlying distribution is one of the inherent premises of semi-supervised learning. In addition, applying other deep unsupervised learning methods, especially generative models, such as variational autoencoder \cite{kingma2013auto} and generative adversarial network \cite{goodfellow2014generative}, for semi-supervised learning requires further research.

\small

\section*{Acknowledgments}
This work is conducted for a course titled `The Principles of Deep Learning,' delivered by Professor Namhoon Lee, Ulsan National Institute of Science and Technology (UNIST). The author is very grateful for his helpful feedback on the draft. This work was chosen as the best paper runner-up in the same course.

\bibliographystyle{unsrt}  
\bibliography{Gyeongho_Kim.bib}

\end{document}